# Entities as topic labels:
# Improving topic interpretability and evaluability
# combining Entity Linking and Labeled LDA

Federico Nanni
Data and Web Science Group
University of Mannheim
federico@informatik.uni-mannheim.de

Pablo Ruiz Fabo
LATTICE Lab (ENS, CNRS, Paris 3)
PSL Research Univ, Sorbonne Paris Cité
pablo.ruiz.fabo@ens.fr

**Introduction**

Humanities scholars have experimented with the potential of different text mining techniques for exploring large corpora, from co-occurrence-based methods to sequence-labeling algorithms (e.g. Named entity recognition). LDA topic modeling (Blei et al., 2003) has become one of the most employed approaches (Meeks and Weingart, 2012). Scholars have often remarked its potential for distant reading analyses (Milligan, 2012) and have assessed its reliability by, for example, using it for examining already well-known historical facts (Au Yeung, 2011). However, researchers have observed that topic modelling results are usually difficult to interpret (Schmidt, 2012). This limits the possibilities to evaluate topic modeling outputs (Chang et al., 2009).

In order to create a corpus exploration method providing topics that are easier to interpret than standard LDA topic models, we propose combining two techniques called Entity linking and Labeled LDA; we are not aware of literature combining these two techniques in the way we describe. Our method identifies in an ontology a series of descriptive labels for each document in a corpus. Then it generates a specific topic for each label. Having a direct relation between topics and labels makes interpretation easier; using an ontology as background knowledge limits label ambiguity. As our topics are described with a limited number of clear-cut labels, they promote interpretability, and this may help quantitative evaluation.

We illustrate the potential of the approach by applying it to define the most relevant topics addressed by each party in the European Parliament's fifth term (1999-2004).

The structure of the abstract is as follows: We first describe the basic technologies considered. We then describe our approach combining Entity Linking and Labeled LDA. Based on the European Parliament corpus[1] (Koehn, 2005), we show how the results of the combined approach are easier to interpret or evaluate than results for Standard LDA.

---

[1] http://www.statmt.org/europarl/

**Basic technologies**

**1 - Entity Linking**

Entity linking (Rao et al., 2013) tags textual mentions with an entity from a knowledge base like DBpedia (Auer et al., 2007). Mentions can be ambiguous, and the challenge is to choose the entity that most closely reflects the sense of the mention in context. For instance, in the expression *Clinton Sanders debate*, *Clinton* is more likely to refer to DBpedia entity *Hillary_Clinton* than to *Bill_Clinton*. However, in the expression *Clinton vs. Bush debate*, the mention *Clinton* is more likely to refer to *Bill_Clinton*. An entity linking tool is able to disambiguate mentions taking into account their context, among other factors.

**2 - LDA Topic Modeling**

Topic modeling is arguably the most popular text mining technique in digital humanities (Brauer and Fridlund, 2013). It addresses a common research need, as it can identify the most important topics in a collection of documents, and how these topics are distributed across the documents in the collection. The method's unsupervised nature makes it attractive for large corpora.
However, topic modeling does not always yield satisfactory results. The topics obtained are usually difficult to interpret (Schmidt, 2012, among others). Each topic is presented as a list of words. It generally depends on the intuitions of the researcher how to interpret these tokens in order to propose concepts or issues that these lists of words represent.

**3- Labeled LDA**

An extension of LDA is Labeled LDA (Ramage et al., 2009). If each document in a corpus is described by a set of tags (e.g. a newspaper archive with articles tagged for areas like 'economics', 'foreign policy', etc.), labeled LDA will identify the relation between LDA topics, documents and tags, and the output will consist of a list of *labeled topics*.

**Our approach**

Labeled LDA has shown its potential for fine grained topic modeling (e.g. Zirn and Stuckenschmidt, 2014). The method requires a corpus where documents are annotated with tags describing their content. Several methods can be applied to automatically generating tags, e.g. keyphrase-extraction (Kim et al. 2010). Our source for tags is Entity linking. Since entity linking provides a unique label for sets of topically-related expressions across a corpus' documents, it can help researchers get an overview of different concepts present in the corpus, even if the concepts are conveyed by different expressions in different documents.

Our first step is identifying potential topic labels via entity linking. Linked entities were obtained with DBpedia Spotlight (Mendes et al., 2011). Spotlight disambiguates against DBpedia, outputting a confidence value for each annotation.[2] Annotations whose confidence was below 0.1 were filtered out. We also removed too general or too frequent entities (e.g. *Country* or *European_Union*)

We then rank entities' relevance per document with tf-idf, which promotes entities that are salient in a specific subset of corpus documents rather than frequent overall in the corpus. Finally, we select the top five entities per document as per tf-idf. These five entities are used as labels to identify, with Labeled LDA, the distribution of labeled topics in the corpus.

**Experiments and Results**

Using the Stanford Topic Modeling Toolbox[3], we performed both Standard LDA (k=300) and Labeled LDA (with 5 labels)[4] on speech transcripts for the 125 parties at the European Parliament (1999-2004 session). The corpus contains 125 documents, representing one party each. Documents were tokenized and lemmatised; stopwords were removed. DBpedia entities were detected with Spotlight and ranked by tf-idf, as described above.

We present the outputs of labeled LDA with entity labels (EL_LDA) for three parties, compared to both Standard LDA and to the top-ranked entities for each party (by tf-idf). In each case, we show topics with relevance above 10%. Results for the remaining parties are available online.[5]

---

[2] Spotlight outperforms other systems when corpus entities often correspond to common-noun mentions like *democracy*, vs. proper-noun mentions (e.g. *Greenpeace*). See Cornolti et al., 2013 and Usbeck et al., 2015.
[3] http://nlp.stanford.edu/software/tmt/tmt-0.4/
[4] Each document (party) is labeled with 5 entities. Some entities are shared across parties. For the 125 parties, this gives 300 distinct labels. This corresponds to k=300 topics in Standard LDA.
[5] https://sites.google.com/site/entitylabeledlda

## Les Verts (France)

| Only Entities - TFIDF ranked | Standard LDA | EL_LDA |
|---|---|---|
| Developing country<br>Consumer<br>Genetically modified org.<br>Development aid<br>Biodiversity | 20%, "political term development case economic community level amendment citizen possible public question market order doe national matter regard situation"<br><br>20%, "gentleman order development lady human greens freedom food asylum citizen fundamental transport directive environment programme resource respect nuclear democracy disaster"<br><br>15%, "economic sustainable developing environmental energy local fishing investment farmer research water production consumer particularly farming oil fishery condition development agriculture"<br><br>10%, "environment amendment public agreement ensure human health directive product safety want long citizen information programme waste vote consumer industry law" | Consumer, 47%<br>Genetically modified organism, 34%<br>Development aid 14% |

## Conservative Party (UK)

| Entities - TFIDF ranked | Standard LDA | EL_LDA |
|---|---|---|
| United Kingdom<br>Conservatism<br>Industry<br>Business<br>British People | 31%: "house, british, want, colleague, amendment, market, industry, united, know, business, going, hope, government, come, rapporteur, said, kingdom"<br><br>14%: ""government, ensure, economic, welcome, world, political, believe, future, common, market, directive, health, consumer, want, million, development, public, decision, farmer, food"<br><br>12%: "economic, social, public, market, measure, situation, financial, level, national, given, service, order, doe, term, community, mean, rapporteur, decision, increase, particularly" | Industry: 35%<br>Business: 34%<br>United Kingdom: 25% |

**Partido Nacionalista Vasco**

| Only Entities - TFIDF ranked | Standard LDA | EL LDA |
|---|---|---|
| Basque Country<br>Basque people<br>Spain<br>Nationalism<br>Terrorism | 100%, "glossed persecute inquisition underscored ulla universe exasperated unquestionable amass ddt condoned estoril cannes deceptive reappearance predominates reclassify corrects hauled remotest" | Basque People, 100% |

**Discussion**

Labeled LDA combines the strengths of Entity Linking and standard LDA. Entity Linking provides clear labels, but no notion of the proportion of the document that is related to the entity. Standard LDA's relevance scores do provide an estimate to what an extent the topic is relevant for the document, but the topics are not expressed with clear labels. Labeled LDA provides both clear labels, and a quantification of the extent to which the label covers the document's content.

An advantage of Labeled LDA over Standard LDA is topic interpretability. Consider the UK Conservative Party's topics. In each standard LDA topic, there are words related to the concepts of *Industry* and *Business* in general, and some words related to the UK appear on the first topic. However, in each topic, some other words (e.g. *government, directive, decision, measure, health, consumer*) are related to other concepts, like perhaps *Legislation* or *Social policy*. A researcher trying to understand the standard LDA topics is faced with choosing which lexical areas are most representative of each topic: is it the ones related to *Industry*, *Business*, and the UK, or is it the other ones? The clear-cut labels from Labeled LDA are more interpretable than a collection of words representing a topic.

The Labeled LDA topics may be more or less correct, just like Standard LDA topics. But we find it easier to evaluate a topic via questions like "is this document about *Industry*, *Business* and *the UK*, in the proportions indicated by our outputs?" than via questions like "is this document about issues like *house, british, amendment, market, industry, government,* (and so on for the remaining topics)"?

The topics for French party Les Verts illustrate Labeled LDA's strengths further. Most of the Standard LDA topics contain some words indicative of the party's concerns (e.g. *environment* or *development*). However, it is not easy to point out which specific issues the party addresses. In Labeled LDA, concrete issues come out, like *Genetically modified organism*.

Topic label *Development aid* shows a challenge with entity linking as a source of labels. Occurrences of the word *development* have been disambiguated towards the entity *Development_aid*, whereas the correct entity is likely *Sustainable_development*. These errors do not undermine the method's usefulness. Efficient ways to filter out such errors exist; this is conceptually similar to removing irrelevant words from Standard LDA topics. However, we need to be aware of and address this challenge.

Regarding Partido Nacionalista Vasco (Basque Nationalist Party), the Standard LDA topic misses the word *basque*, which is essential to this party. Labeled LDA identifies *Basque people* as a dominant concept in this party's interventions.

**Outlook**

Our method performs Labeled LDA using Entity Linking outputs as labels. Its main advantage is providing a specific label for each topic, that improves topic interpretability, and can simplify human evaluation of topic models.

More evaluation is needed to fully assess the approach. We will consider two possible complementary evaluations: first, a crowdsourced task where participants evaluate the coherence of Labeled LDA topics with the corpus documents. Second, an assessment of our topics by political science experts. We're mostly interested in evaluating the approach for diachronic comparisons.

**Acknowledgements**
Pablo Ruiz was supported by a PhD scholarship from Région Île-de-France.

**References**

Au Yeung, Ching-man, and Adam Jatowt. "Studying how the past is remembered: towards computational history through large scale text mining."Proceedings of the 20th ACM international conference on Information and knowledge management. ACM, 2011.

Auer, Sören, et al. "Dbpedia: A nucleus for a web of open data". The Semantic Web. Springer Berlin Heidelberg, 2007.

Blei, David M., Andrew Y. Ng, and Michael I. Jordan. "Latent dirichlet allocation." the Journal of machine Learning research 3 (2003): 993-1022.

Brauer, René, and Mats Fridlund. "Historicizing Topic Models, A distant reading of topic modeling texts within historical studies." International Conference on Cultural Research in the context of "Digital Humanities", St. Petersburg: Russian State Herzen University. 2013.

Chang, Jonathan, et al. "Reading tea leaves: How humans interpret topic models." Advances in neural information processing systems. 2009.

Cornolti, Marco, Paolo Ferragina, and Massimiliano Ciaramita. "A framework for benchmarking entity-annotation systems." Proceedings of the 22nd international conference on World Wide Web. International World Wide Web Conferences Steering Committee, 2013.

Kim, Su Nam, Olena Medelyan, Min-Yen Kan, and Timothy Baldwin. (2010). "Semeval-2010 Task 5: Automatic Keyphrase Extraction from Scientific Articles." In Proceedings of the 5th International Workshop on Semantic Evaluation, 21–26. Association for Computational Linguistics.

Koehn, Philipp. "Europarl: A parallel corpus for statistical machine translation."MT summit. Vol. 5. 2005.

Mendes, Pablo N., et al. "DBpedia spotlight: shedding light on the web of documents." Proceedings of the 7th International Conference on Semantic Systems. ACM, 2011.

Meeks, Elijah, and S. Weingart. "The digital humanities contribution to topic modeling." Journal of Digital Humanities 2.1 (2012): 1-6.

Milligan, Ian. "Mining the 'Internet Graveyard': Rethinking the Historians' Toolkit." Journal of the Canadian Historical Association/Revue de la Société historique du Canada 23.2 (2012): 21-64.

Ramage, Daniel, et al. "Labeled LDA: A supervised topic model for credit attribution in multi-labeled corpora." Proceedings of the 2009 Conference on Empirical Methods in Natural Language Processing: Volume 1-Volume 1. Association for Computational Linguistics, 2009.

Rao, Delip, Paul McNamee, and Mark Dredze. "Entity linking: Finding extracted entities in a knowledge base." Multi-source, multilingual information extraction and summarization. Springer Berlin Heidelberg, 2013. 93-115.

Salton, Gerard, Edward A. Fox and Harry Wu. (1983). Extended Boolean information retrieval. Communications of the ACM 26(11): 1022-1036

Schmidt, Benjamin M. "Words alone: Dismantling topic models in the humanities." Journal of Digital Humanities 2.1 (2012b): 49-65.

Ricardo Usbeck, Michael Röder, Axel-Cyrille Ngonga Ngomo. (2015). Evaluating Entity Annotators using GERBIL. Proceedings of ESWC, the 12th European Semantic Web Conference.

Zirn, Cäcilia, and Heiner Stuckenschmidt. "Multidimensional topic analysis in political texts." Data & Knowledge Engineering 90 (2014): 38-53.